\documentclass{article}

\usepackage{arxiv}

\usepackage[utf8]{inputenc} % allow utf-8 input
\usepackage[T1]{fontenc}    % use 8-bit T1 fonts
\usepackage{hyperref}       % hyperlinks
\usepackage{url}            % simple URL typesetting
\usepackage{booktabs}       % professional-quality tables
\usepackage{amsfonts}       % blackboard math symbols
\usepackage{nicefrac}       % compact symbols for 1/2, etc.
\usepackage{microtype}      % microtypography
\usepackage{lipsum}
\usepackage{graphicx}
\usepackage{tabularx}

\usepackage{color-edits}
\addauthor[Rayid]{rg}{blue}
\addauthor[Kit]{kr}{red}
\addauthor[Carly]{cj}{RedViolet}

\title{Breaking the Cycle of Incarceration With Targeted Mental Health Outreach: A Case Study in Machine Learning for Public Policy}

\author{
  Kit T.~Rodolfa \\
  Carnegie Mellon University\\
  \texttt{krodolfa@cmu.edu} \\
   \And
  Erika~Salomon \\
  Carnegie Mellon University \\
  \texttt{ecsalomon@gmail.com} \\
   \And
  Jin~Yao \\
  Johnson County, KS \\
  \texttt{Jin.Yao@jocogov.org} \\
   \And
  Steve~Yoder \\
  Johnson County, KS \\
  \texttt{Steve.Yoder@jocogov.org} \\
   \And
  Robert~Sullivan \\
  Johnson County, KS \\
  \texttt{Robert.Sullivan@jocogov.org} \\
   \And
  Kevin~McGuire \\
  Johnson County, KS \\
  \texttt{Kevin.McGuire@jocogov.org} \\
   \And
  Allie~Dickinson \\
  Johnson County, KS \\
  \texttt{Allie.Dickinson@jocogov.org} \\
   \And
  Rob~MacDougall \\
  Johnson County, KS \\
  \texttt{Rob.MacDougall@jocogov.org} \\
   \And
  Brian~Seidler \\
  Johnson County, KS \\
  \texttt{Brian.Seidler@jocogov.org} \\
   \And
  Christina~Sung \\
  University of Chicago \\
  \texttt{christina.sung@gmail.com} \\
   \And
  Claire~Herdeman \\
  University of Chicago \\
  \texttt{cherdeman@gmail.com} \\
   \And
  Rayid~Ghani \\
  Carnegie Mellon University \\
  \texttt{rayid@cmu.edu} \\
}

\begin{document}

\maketitle

\begin{abstract}
Many incarcerated individuals face significant and complex challenges, including mental illness, substance dependence, and homelessness, yet jails and prisons are often poorly equipped to address these needs. With little support from the existing criminal justice system, these needs can remain untreated and worsen, often leading to further offenses and a cycle of incarceration with adverse outcomes both for the individual and for public safety, with particularly large impacts on communities of color that continue to widen the already extensive racial disparities in criminal justice outcomes. Responding to these failures, a growing number of criminal justice stakeholders are seeking to break this cycle through innovative approaches such as community-driven and alternative approaches to policing, mentoring, community building, restorative justice, pretrial diversion, holistic defense, and social service connections. Here we report on a collaboration between Johnson County, Kansas, and Carnegie Mellon University to perform targeted, proactive mental health outreach in an effort to reduce reincarceration rates. 
%individuals at risk for returning to jail within a year. 
This paper describes the data used, our predictive modeling approach and results, as well as the design and analysis of a field trial conducted to confirm our model's predictive power, evaluate the impact of this targeted outreach, and understand at what level of reincarceration risk outreach might be most effective. Through this trial, we find that our model is highly predictive of new jail bookings, with more than half of individuals in the trial's highest-risk group returning to jail in the following year. Outreach was most effective among these highest-risk individuals, with impacts on mental health utilization, EMS dispatches, and criminal justice involvement.
\end{abstract}

% keywords can be removed

\section{Introduction}
In the United States, individuals who have complex sets of needs, including mental health conditions, long-term unemployment, substance abuse, and homelessness, can many times find themselves in repeat interactions with the criminal justice system. For this vulnerable population, the combined challenges they face may increase their likelihood of short-term detainment or long-term imprisonment as a result of these interactions, but jails and prisons are poorly suited to addressing these needs, often leading to worsening challenges and further incarceration \cite{Hamilton2010PeopleSystem}. On the whole, more than half of the population of incarcerated individuals held in the country's prisons and jails each year face current or recent mental illness. In local jails, 55\% of individuals met criteria for substance abuse or dependence, 64\% struggled with mental health issues, and the jailed population was 10 times as likely to have experienced homelessness in the previous year as the general population \cite{James2006MentalInmates,USHUD2017Annual2}.

These statistics paint a grim picture of the circumstances faced by the incarcerated, which are exacerbated further by often limited options for these individuals to receive care. Over the last half-century, the per capita number of psychiatric hospital beds has fallen by an order of magnitude, and by 2005, jails and prisons housed three times as many individuals with serious mental illnesses as hospitals \cite{FullerTorrey2010MoreStates}. In some cases, an individual's initial or primary point of contact with social services might be in the context of detainment or confinement, yet this context is particularly poorly suited to resolving this complex array of needs. In the absence of appropriate treatment or care, an individual's condition may degrade further, leading to a sizable group of people cycling in and out of the nation's jails and prisons \cite{Stone1997TherapeuticPolicy,Kondo2000TherapeuticOffenders.,Kutcher2009ProblemsSystem}. Further reinforcing this cycle is a system of what Demleitner \cite{Demleitner2002CollateralOffenders} terms ``collateral restrictions'' such as restrictions on drawing social benefits or gaining employment for convicted individuals, especially those with drug offenses or facing substance abuse issues. This situation is beneficial to neither the individuals caught in the system nor society at large, reflecting a failure to improve public safety while expending billions of dollars each year on the costs of incarceration. Recidivism rates among individuals with mental illness are as high as 70\% in some jurisdictions \cite{Ventura1998CaseJail}, and surveys of practitioners and stakeholders in both the mental health and criminal justice fields have found widespread, bipartisan dissatisfaction with the state of mental health care and resources for those involved in the criminal justice system \cite{Thompson2003CriminalIllness}.

Furthermore, the burden of this relationship between complex needs and criminal justice involvement falls disproportionately on Black and Hispanic individuals, who face much higher rates of homelessness \cite{USHUD2018Annual1}, are more than twice as likely to live in poverty \cite{Creamer2020Poverty2019}, and face significant barriers in access to and quality of mental health care \cite{McGuire2008NewImplications} relative to their white counterparts. These multifaceted disparities may in turn exacerbate the stark disparities in criminal justice outcomes, where the rate of incarceration of Black individuals is more than six times that of white individuals and minority communities have faced a long history of profiling and over-policing \cite{Hinton2018AnSystem}. The May 2020 murder of George Floyd by Minneapolis police and subsequent growth of the Black Lives Matter movement throughout the following summer has increased the public awareness of the impact of racism in the criminal justice system and sparked a national conversation about reforming policing and prisons \cite{Cohn2020HowMatter,Winter2020VotersPolice,Peeples2020WhatWork}. The work we present here contributes to this conversation by focusing specifically on disrupting the interaction between mental health needs and criminal justice involvement, which we believe could be one valuable element of a broad range of reforms aimed at improving outcomes and fairness in the US criminal justice system.

\subsection{Intervening to Break the Cycle}
In recent years, a growing number of communities have begun to explore innovative approaches to break this adverse cycle for individuals facing complex needs, seeking to reduce overcrowding in jails and the high costs of incarceration \cite{Steadman2005AssessingDisorders,Thompson2003CriminalIllness}. Ranging from restorative justice to pretrial diversion to social service connections, the scope, design, and population served by these programs vary considerably. For instance, some focus on community services or facilitated meetings between victims and offenders while others seek to directly address mental illness and substance dependence issues through treatment or other resources. Several studies have sought to evaluate the impact of these interventions \cite{Steadman2005AssessingDisorders,Hartford2007PretrialIllness,Cosden2003EvaluationTreatment,Lamb1996CourtOffenders}, indicating that they can be successful in both reducing time spent incarcerated and increasing social service utilization without increasing public safety risks (evidence around the financial costs and savings of these programs, however, is more mixed \cite{Cowell2004TheAbuse,Steadman2005AssessingDisorders}). 

A number of recent reviews have focused specifically on the treatment and intervention points for mentally ill individuals involved with the criminal justice system \cite{Kim2015TheSystem,Kennedy-Hendricks2016ImprovingIllness,Sarteschi2013MentallySynthesis}, describing considerable variation across jurisdictions in approaches to addressing the needs of this population, generally falling short of meeting the high levels of need across the system. Nevertheless, these studies do highlight some promising approaches that may improve both mental health and criminal justice outcomes, but highlight the rarity of strong empirical evidence based on rigorous evaluation such as randomized trials. Among the practices focused on this population, two that have found relatively wide adoption are mental health courts and Forensic Assertive Community Treatment (FACT) programs. 

Mental health courts serve as an alternative route for individuals facing charges who are identified as having significant mental health challenges, and focus on problem-solving solutions including treatment and social service diversions. Use of these special courts has grown rapidly recently, from 75 in 2003 to more than 350 by 2016 \cite{Kennedy-Hendricks2016ImprovingIllness}, yet relatively few of these programs have been the subject of rigorous evaluations \cite{Kim2015TheSystem}. Sarteschi and colleagues \cite{Sarteschi2011AssessingReview} conducted a meta-analysis of published evaluations of Mental Health Courts and found these programs could be effective at reducing future contacts with the criminal justice system, but noted that the strength of their conclusions were limited by the methodological strength of the underlying quasi-experiment evaluations they studied. Likewise, in \cite{Kim2015TheSystem}, the authors note that they were aware of only two randomized trials which provided stronger evidence for improvements of clinical outcomes but were less conclusive for criminal justice outcomes.

FACT programs extend upon the Assertive Community Treatment model developed in the 1970s as a model of high-touch, individualized treatment for individuals with severe and persistent mental illness. Focused on individuals with a history of criminal justice involvement, this model seeks to prevent future incarceration as well as improve clinical outcomes. As is the case with mental health courts, most evaluations of FACT programs have been quasi-experimental, but provide promising suggestive evidence for improvements in criminal justice outcomes \cite{Marquant2016ForensicLiterature}. Notably, a few randomized trials provide some additional strength to these conclusions. Cusack and colleagues compared outcomes for 134 individuals randomized into either FACT or traditional treatment options, finding the FACT-treated individuals had fewer jail bookings, more outpatient mental health contact, and fewer hospital days \cite{Cusack2010CriminalTrial}. More recently, Lamberti and colleauges \cite{Lamberti2017AModel} conducted a randomized evaluation of the Rochester, NY FACT program, finding strong effects on convictions, days spent in jail, and hospital days.

\subsection{Existing Approaches in Johnson County, KS}
The current work describes a multi-year collaboration between Johnson County, KS and Carnegie Mellon University,\footnote{The collaboration originally began in 2016 through the Data Science for Social Good program at the University of Chicago, but the researchers involved have since moved to Carnegie Mellon University} focused on reducing incarceration for individuals with mental health needs and past criminal justice involvement. Johnson County is the most populous county in the state of Kansas, with many of its 20 municipalities comprising the suburbs of Kansas City, MO. While Johnson County has a number of programs that help different government systems share resources and identify needs in cross-functional ways, individuals who need assistance are often identified in reactive ways, after experiencing a mental health crisis or interacting with the criminal justice system. 

Currently, individuals with complex mental health needs typically become involved with the Johnson County Mental Health Center (JCMHC) through one of three avenues: First, they may be referred to the center, which depends on the individual in need of assistance and on their environment. Second, their needs may be identified by co-responders in police departments and the Emergency Medical Services (EMS). These mental health professionals interact on cases with mental service needs to improve outcomes for people, largely by following up after police or EMS contact. And, third, through screening instruments among individuals entering jail, such as the Brief Jail Mental Health Screen (BJMHS), used to identify individuals in need of mental health assistance for follow-up upon their release, or the Level of Service Inventory - Revised (LSI-R), used to assess a range of needs and risk factors related to criminal justice involvement. Recently, JCMHC staff has begun serving the incarcerated population directly as well, with staff offering counseling and services to current inmates.

\subsection{Project Scope and Goals}
The key goal of the current work is to supplement Johnson County's existing, reactive efforts with a more proactive mechanism for identifying individuals who may be helped by outreach from JCMHC's Mobile Crisis Response Team (MCRT). Because the MCRT team has limited capacity for outreach, they wanted to focus their resources on individuals with high needs who are at risk of experiencing a new jail booking with the hope that proactive outreach might help resolve their mental health issues and thereby reduce their incarceration risk. In order to prioritize individuals for proactive mental health outreach, JCMHC enlisted the help of the Data Science for Social Good program in 2016 to develop a predictive model of reincarceration risk. This effort brought together data from Johnson County’s criminal justice, emergency medical, and mental health systems, and preliminary results of the modeling work were described in \cite{Bauman2018ReducingInterventions}. This work expands upon those initial results, presenting updated modeling results as well as the design and analysis of a field trial conducted to confirm our model's predictive power, evaluate the impact of this targeted outreach, and understand at what level of reincarceration risk outreach might be most effective. Through this trial, we find that our model is highly predictive of new jail bookings, with more than half of individuals in the trial's highest-risk group returning to jail in the following year. Outreach was most effective among individuals with the highest risk, with impacts on mental health utilization, EMS dispatches, and criminal justice involvement.
\rgcomment{add equity and ethics goals - efficiency, cost savings, vs improving outcomes and helping people}

\section{Data and Modeling}

Through our partnership, Johnson County provided administrative data from their mental health center, jail system, police arrests from the county's two largest communities, county department of health, and ambulance runs. Each month, updated extracts of these datasets were obtained by the Johnson County Department of Technology and Innovation (DTI) where record linkage was performed using a set of heuristics on identifiable information from each system (such as names, dates of birth, and social security numbers). The linked records were assigned unique identifiers and then de-identified data was transferred to our team for the modeling work. Even though efforts had been made to obscure the identities of individuals in these data, there is always a risk that it could be re-identified, particularly given the large number and types of features available about each individual. As such, the data was treated as highly sensitive and all the work described here took place under strict data use agreements and in secure server environments.

The data included information about any individual who had interacted with any of these systems (police, ambulance, county jail, health department, or mental health center), regardless of whether they were current or past residents of Johnson County, with reliable historical data dating back to at least 2010 (and earlier from some of the sources). In the most recent extract as of this writing (January 2021), the data included information about 1,475,890 distinct individuals and 9,474,518 events. Because this outreach program aims to prevent returns to jail, modeling focused on a cohort of individuals who had been released from jail in the past three years and were not currently in jail (approximately 27,000 individuals meet this cohort definition at any given point in time). This cohort was used for model development and selection, but then further narrowed to Johnson County residents with any history of mental health needs as a target population for outreach, reflecting JCMHC's criteria for inclusion (approximately 8,000 individuals meet this subset definition at any given point in time). History of mental health needs was identified based on contacts with JCMHC, inmate screening instruments (specifically the Level of Service Inventory - Revised, LSI-R, and Brief Jail Mental Health Screen, BJMHS), or when responder notes from ambulance runs mentioned substance, behavioral, or psychiatric issues.

\begin{figure}
  \centering
  \includegraphics[width=\linewidth]{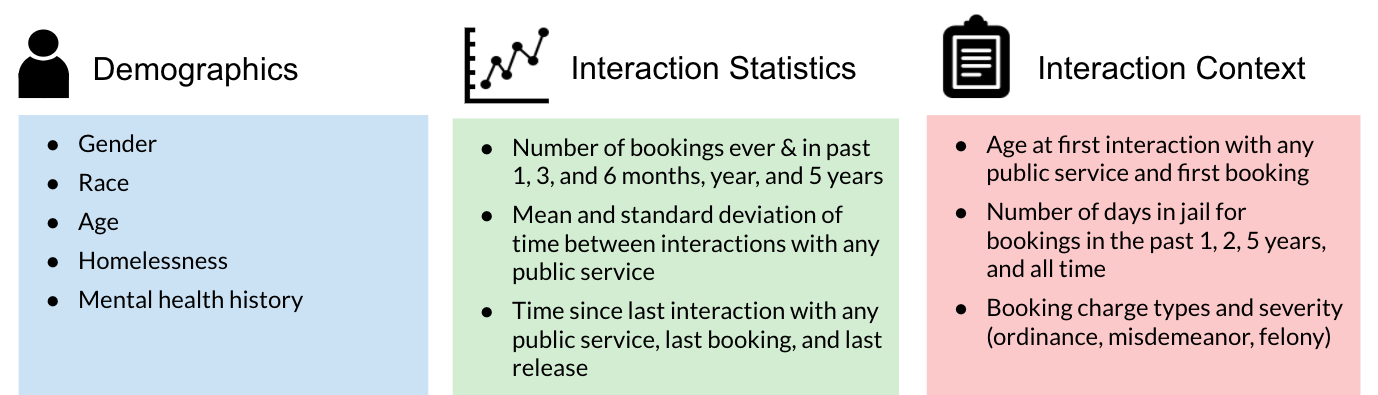}
  \caption{Descriptions of a sample of features used for modeling.}
  \label{fig:model_features}
\end{figure}

This raw input data was structured into features to include information about an individual's demographics, history of interactions with the systems included in the data, and context around those interactions (Figure \ref{fig:model_features} highlights some examples of these features). The modeling task focused on predicting whether an individual in the cohort would return to jail within one year, specifically focusing on new jail bookings types ``on view'' (when someone is arrested by an officer on patrol), ``warrant'' (when someone is arrested with an arrest warrant), or ``bench warrant'' (generally issued when someone fails to appear for a court date). Although the data also includes post-conviction jail bookings that might stem from a prior arrest or ongoing trial, these were excluded as outcome labels in order to focus on bookings for new, potentially preventable interactions with the criminal justice system.

We used a strategy of inter-temporal cross-validation (as described in \cite{Roberts2017Cross-validationStructure} and \cite{Ye2019UsingCity}) to ensure that model evaluation and selection was done in a manner that reflected performance on novel data while guarding against ``leakage'' of information from the future affecting past results. Each temporal validation split simulates standing at a given point in time, with only information prior to that date available for model training and labels drawn from the following year. Reflecting the monthly cadence of data updates and JCMHC outreach planning, we created temporal validation splits for each month from January 2013 through May 2018 (with the final year's worth of data through May 2019, when the modeling was performed, reserved for the last validation split).

All machine learning models, including feature engineering, model training, and performance evaluation were run using our open-source \verb|python| ML pipeline package, \verb|triage|. Machine learning methods used are from \verb|sklearn| (a python package) or \verb|catwalk| (a component of \verb|triage| for baselines methods as well as \verb|ScaledLogisticRegression|, which wraps the \verb|sklearn| logistic regression to ensure input features as scaled between 0 and 1). The modeling grid used for this work is described in Tables \ref{tab:model_grid}, reflecting the space of model types and hyperparameters explored here. For each estimator in the tables, the grid search considered reflects the full cross-product of the hyperparameter values specified. We also initially considered (including in the early work reported in \cite{Bauman2018ReducingInterventions}) K nearest neighbor, decision tree, extra trees, and AdaBoost classifiers, but didn't find gains from using these methods and narrowed our modeling grid for subsequent work here.

\begin{table}[htb!]
 \caption{Model Hyperparameter Grid}
  \centering
  \begin{tabularx}{\linewidth}{ l l X }
    % \toprule
    \hline
    Estimator     & Hyperparamter & Grid Search Values  \\
    % \midrule
    \hline
    \verb|RandomForestClassifier| & max\_features & sqrt \\
    & criterion & gini, entropy \\
    & n\_estimators & 100, 1000, 5000 \\
    & min\_samples\_split & 10, 25, 100 \\
    & max\_depth & 5, 10, 50 \\
    % \midrule
    \hline
    \verb|ScaledLogisticRegression| & penalty & l1, l2 \\
    & C & 0.001, 0.1, 1, 10 \\
    % \midrule
    \hline
    \verb|PercentileRankOneFeature| & feature & Jail bookings last 1 or 5 years \\
    % \midrule
    \hline
    \verb|SimpleThresholder| & rules & Any mental health history, \\
    & & Jail releases in last: 1, 3, 6 months, or 1 year \\
    & logical\_operator & and \\
    % \bottomrule
    \hline
  \end{tabularx}
  \label{tab:model_grid}
\end{table}

\section{Modeling Results}

\begin{figure}
  \centering
  \includegraphics[width=0.57\linewidth]{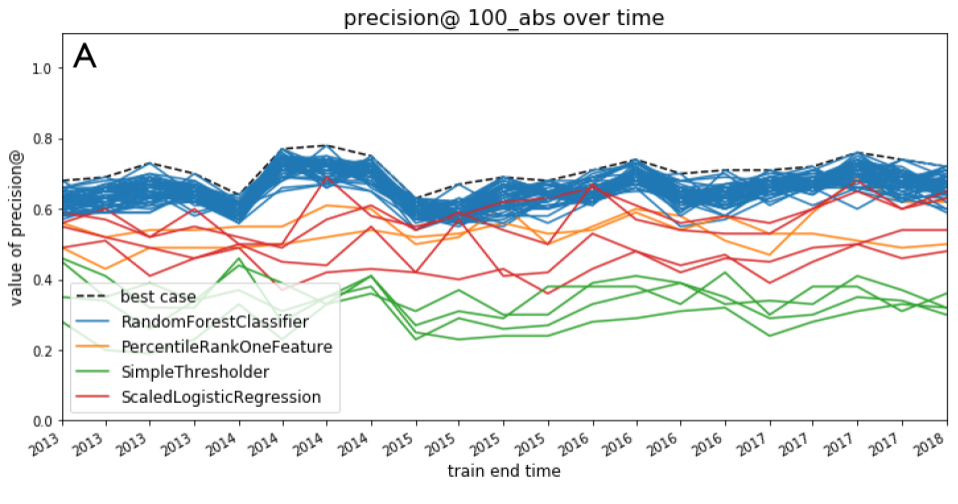}
  \includegraphics[width=0.42\linewidth]{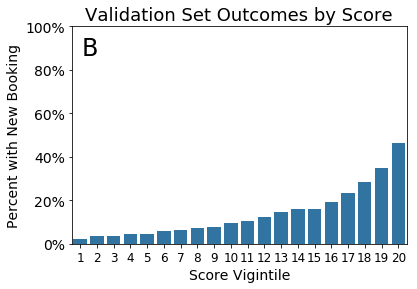}
  \includegraphics[width=\linewidth]{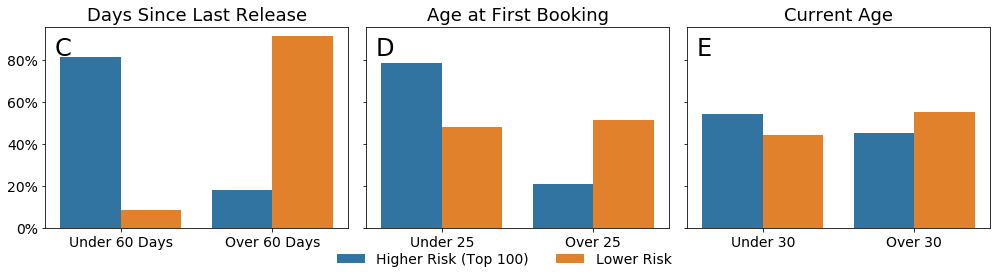}
  \caption{(A) Temporal validation performance for the full grid of model specifications across all hold-out validation sets, showing the precision (Positive Predictive Value) among the 100 highest-risk individuals on each validation cohort. (B) Performance of the model selected for field trial across score vigintiles on the latest validation set. (C-E) Comparison of feature distributions between the 100 highest-risk individuals (blue bars) and all other individuals (orange bars) for the number of days since the most recent release from jail (C), age at first booking (D), and current age (E).}
  \label{fig:model_results}
\end{figure}

\subsection{Model Selection and Performance}
Reflecting MCRT's capacity for proactive outreach of 100 individuals per month, we evaluated our machine learning models on the basis of precision among the top 100 (precision@100). Here, precision (also know as positive predictive value) can be seen as a measure of efficiency of the model's recommendations: of the 100 individuals who would be selected for outreach, what fraction of them actually do experience a new jail booking in the following year? Figure \ref{fig:model_results}A shows the performance on this metric for each model specification considered across each of the temporal validation splits described above. Here, each color indicates a different class of machine learning model and each line a given set of model hyperparameters. Here, random forest models consistently out-performed other classes of models considered, and we chose a specific random forest specification based on average performance across validation sets as well as how frequently the specification was among the best-performing models across these validation sets. Figure \ref{fig:model_results}B shows the performance of the chosen model specification on the final validation set broken out by model vigintile. The model rank-orders individuals well, with new booking rates decreasing across the vigintiles, ranging from 46\% for the 5\% with the highest model scores to just 2\% for the bottom 5\% of individuals. Among the 100 individuals identified with the highest model risk in this final year of validation data, 82 were booked into jail. This represents a  12\% improvement over a baseline based on the number of bookings in the prior year and 4.8-fold increase over the population prevalence.

\subsection{Characteristics of Model Targets}
Figure \ref{fig:model_results}(C-E) illustrates how the 100 highest-risk individuals compare with the remainder of the population across various attributes. Recent interaction with the criminal justice system was particularly informative to the models, with more than 80\% of these 100 highest-risk individuals having most recently been released from jail no more than 60 days before (compared with fewer than 10\% of other individuals). Age was also an important predictor, both current age (with a majority of the highest-risk individuals under 30), and, more dramatically, age at the time of first jail booking (approximately 3/4 of higher-risk individuals were under 25 at the time of their first booking). This last finding is consistent with other research indicating that early involvement with the criminal justice system is associated with a higher likelihood of future involvement \cite{Pritchard1979StableSummary}.

\subsection{Algorithmic Fairness}
Given both the high stakes involved and the historical disparities in criminal justice outcomes, a key goal of this work is to ensure that the mental health outreach attempts being allocated here are distributed fairly, particularly across racial and ethnic groups. Doing requires mapping the somewhat nebulous concept of ``fairness'' to a concrete and measurable metric that can be applied to the model's recommendations. Researchers in the field of fair machine learning have proposed and enumerated a wide variety of measures that may capture some conceptualization of fairness \cite{Verma2018FairnessExplained,Gajane2018OnLearning}, as well as showing the important conclusion that many of definitions of fairness cannot be satisfied at the same time in the absence of perfect prediction \cite{Chouldechova2017FairInstruments,Kleinberg2017InherentScores}. As such, the choice of an appropriate fairness metric is inherently context-dependent and reflect the goals and values of the stakeholders in a given project. In similar settings where an organization is seeking to allocate a limited set of resources to benefit recipients, we have argued that disparities in recall (also known as sensitivity or true positive rate) is a reasonable conceptualization of fairness \cite{Rodolfa2020CaseInterventions,Rodolfa2020MachineFair}, drawing on the discussion provided by Hardt \cite{Hardt2016EqualityLearning} that describes this metric as a reflection of ``equality of opportunity:'' when only a fraction of individuals with need can receive a benefit, recall measures the fraction of those individuals a program reaches and disparities in recall therefore measure whether individuals with need across different sub-groups have an equal chance of receiving the benefit.

\begin{figure}
  \centering
  \includegraphics[width=0.5\linewidth]{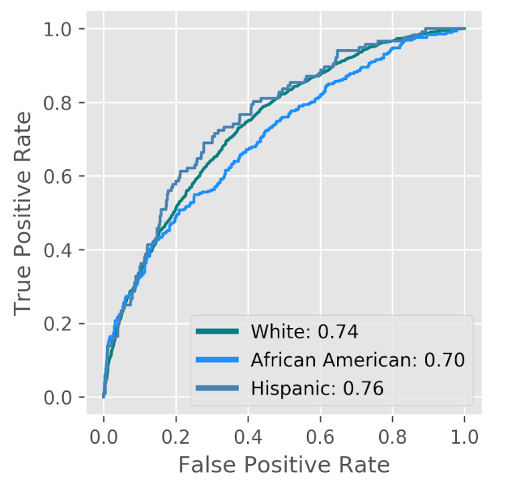}
  \caption{Receiver Operating Characteristic (ROC) Curves by race and ethnicity for the selected model, showing the relationship between false and true positive rates at different thresholds of the model score. Area under the curve (AUC) for each group is shown in the legend.}
  \label{fig:roc_by_race}
\end{figure}

In a recent related work \cite{Rodolfa2020MachineFair}, we considered recall disparities across race and ethnicity among the highest-risk individuals identified by the score developed here. There we found that white individuals who actually did return to jail in the subsequent year were selected by the model for intervention at a 50\% higher rate than non-white individuals who would also return to jail. However, this considerable disparity could be mitigated without loss in overall model precision by choosing recall-equalizing score thresholds across groups. However, one challenge in the present work arises from uncertainty around where in the score distribution outreach might be most effective, as discussed in detail below. Given this uncertainty, we also looked at the Receiver Operating Characteristic (ROC) curves for the selected model, broken out by race and ethnicity, which describe the relationship between false and true positive rates at different score thresholds. The results in Figure \ref{fig:roc_by_race} show similar curves near the top of the risk score, but considerable divergence in the middle and lower ranges of the score, with performance degraded for Black individuals relative to white and Hispanic individuals. Even where the ROC curves overlap considerably, it should also be noted that this is not sufficient to guarantee fairness in model predictions: applying a uniform score threshold across groups may operate at different points on the ROC curves for each group, giving rise to disparities despite superficially-similar curves, as illustrated  by Chouldechova and colleagues in the context of child welfare screening decisions \cite{Chouldechova2018ADecisions}. In light of these results, the implications of deploying this model on the fairness of the mental health outreach efforts -- and appropriate mitigation strategies to reduce disparities -- will depend on the details of how the risk score should best be used to inform the outreach. In the remainder of the work described here, we explore those details in the context of a field trial.

\section{Field Trial Design}

\subsection{Design Overview and Motivation}
Having selected a model and validated it on historical data, we next sought to assess both how well this model generalizes onto unseen future data and how best JCMHC could use it to prevent adverse outcomes in the population. In short, the trial design proposed here aims to answer two questions: First, assessing the accuracy of the model’s predictions. And, second, understanding what risk profile might be most helped by outreach from the JCMHC.

To answer both of these questions concurrently while working within the limits of JCMHC’s ability to do additional outreach beyond their current programs, the trial design randomly assigned individuals into two groups: one that will be eligible to receive this additional outreach and a second that will receive only the current outreach. In the first group, we chose individuals stratified across tiers of predicted risk for outreach, using the second group as a baseline to compare them against. Each of these aspects of the study design is described in more detail below.

\subsection{Research Questions}
While current avenues for finding individuals for outreach allows Johnson County to identify many individuals who may be helped by their mental health programs around the time of a period of incarceration, the goal of this research is to see if predictive modeling can extend those efforts to identify additional individuals for outreach in a more proactive fashion before they return to jail. When applying an intervention like personal outreach to reduce incarceration, the goal is to direct resources to those for whom the intervention is most likely to help prevent future bookings. Because the model predicts the likelihood of reincarceration rather than the likelihood to be helped by the intervention, evaluating the utility of this model entails answering two questions:

\begin{enumerate}
    \item How well do the model and intervention identify risks and improve outcomes for individuals receiving outreach?
    
    \item Do intervention efforts have a greater effect on those who the model assesses as most at risk to return to jail, moderately at risk, or less at risk?
\end{enumerate}

In evaluating the impact of the model-targeted mental health outreach, we are concerned with the average effect of implementing the program across the entire population to whom it would be applied. Notably, some people selected for outreach by the model will not receive treatment for reasons out of MCRT’s control (e.g., current contact information cannot be found for them or they refuse contact when attempted), and this will make the program overall less effective. We therefore focus here on estimating intent-to-treat (ITT) effects, namely the estimated impact of being selected by the model for outreach, regardless of whether contact was made, because the cost of program implementation includes those who are not treated. Moreover, disposition data isn't available for many of the contact modalities used here (e.g., voicemail, door hangers, and direct mail).

The second question asks whether the outreach is most effective for those at high, medium, or lower risk of returning to jail. In the absence of historical label data indicating actual responsiveness to outreach\footnote{Even if the outreach program described here were not new, historical outreach data would lack counterfactual information to model responsiveness directly.}, we trained models here against the outcome of returning to jail in the next year. However, it's not clear that individuals at the top of this risk score would necessarily be those most helped by the outreach in terms of preventing future jail incarceration. For instance, it could be the case that the level of need for these individuals might be too high for the program to have an appreciable effect, while individuals with more moderate levels of need could be better assisted through the outreach.

\subsection{Randomization and Risk Tiers}

\begin{figure}
  \centering
  \includegraphics[width=\linewidth]{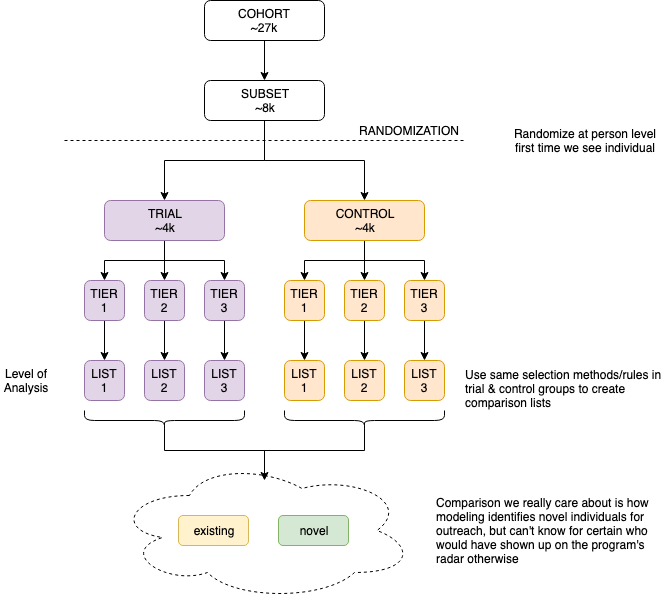}
  \caption{Design of the field trial. Individuals are randomized into either the trial or control arm when they first appear in the data. To assess who is best served by the program, each arm is divided into three tiers based on the predicted risk of reincarceration and interventions are divided across these tiers.}
  \label{fig:trial_design}
\end{figure}

The current trial seeks to address both of these questions concurrently. Each month, the predictive modeling created risk scores for everyone who has had a booking in Johnson County in the last three years and is not currently in jail in Johnson County. From this ranked risk list, 100 people were prioritized for outreach by the Johnson County Mobile Crisis Response Team (MCRT). These 100 people were selected from three different risk tiers, as depicted in Figure \ref{fig:trial_design}. To ensure that all subjects received the same attention, regardless of tier, the MCRT was kept blind to both the risk tier and the ranking within the risk tier. After being included in either an outreach or control list, an individual's outcomes over the following year were assessed.

Note as well that the proactive outreach program being considered here is in addition to existing JCMHC programs, so we are concerned with the impact of this new program relative to the status quo in which some individuals might already receive JCMHC services through other routes. As a result, the mechanism by which the model can help reduce reincarceration is by identifying novel individuals for intervention by the MCRT who would not otherwise be identified for outreach (or, potentially, by identifying these individuals earlier than they would have been contacted otherwise). To assess the potential overlap between the model-targeted population and contacts through existing JCMHC programs, we looked at how many of the 100 highest-risk individuals identified by models as of January 2019 ended up interacting with the mental health center in the following month. Of this group, only 10 of them interacted with JCMHC in the following month, suggesting the capacity for our model to decrease reincarceration is through its impact on the remaining 90\%.

% Given that Johnson County has existing programs that refer people with recent jail bookings to the MCRT, our goal was to identify new referrals who would not otherwise be contacted by the team. In the first month of the trial, we can know who would receive outreach through existing programs because they will have BJMHS scores that indicate referrals or will be visited in jail by the JCMHC. However, some of the people who are contacted through the model-based referral program in the first few months might be people who would otherwise have been referred to the MCRT, just at a later date. As the trial progresses, more and more such people may be referred by the model, complicating the issue of identifying new referrals. We cannot know for a given individual what might have happened to them in the absence of the program, so we need to compare these individuals with a similar population for whom only the existing programs are used to identify individuals for outreach. The proposed design will help us measure the aggregate effect of adding the model-based outreach to the current program. 

Mechanically, the study design involves several steps illustrated in Figure \ref{fig:trial_design}:

\begin{enumerate}
    \item Risk modeling was performed among a cohort of approximately 27,000 people who have had a booking in Johnson County in the last three years and is not currently in a Johnson County jail. This cohort is then subsetted down to the approximately 8,000 individuals relevant to the MCRT’s program based on their most recent address being in Johnson County and on having some indicator of previous mental or behavioral health issue, including substance use (see modeling details above).
    
    \item Among this subset, individuals were randomized once to be included in the trial or control group for the study (individuals existing in the dataset were randomized at the outset of the study, while those newly appearing in the dataset as the study proceeded were randomized when they were first seen). This was done to prevent individuals moving between trial and comparison lists in different months.
    
    \item Each month, risk scores were generated for all individuals in the subset, which were used to assign individuals to three model tiers: high, medium, and low risk (based on their model score).
    \begin{enumerate}
        \item The high risk tier represents the top quintile (20\%) of the model
        \item The medium risk tier represents the middle quintile of the model
        \item The low risk tier represents the bottom two quintiles of the model
    \end{enumerate}
    
    \item Trial and control lists were selected each month with the aim of replicating the process of selecting 100 people in a given tier for intervention, even though only 20 or 40 people were selected from each tier each month. 
    
    \begin{enumerate}
        \item In the high risk tier, we focused on individuals at the top of the score distribution. Because people are randomly assigned to trial and control, we can assume that, on average, 50 of the top 100 in this tier will be in the trial group and the other will be in the control group. To ensure that we selected from the full range of people in the top 100, each trial and control list was randomly sampled from the top 50 trial or control individuals in the tier. In the trial group, 40 of the 50 individuals with the highest modeled risk in tier 1 were provided to the MCRT for outreach.
        
        \item Because scores in the middle and near the bottom of the risk distribution are closer together than at the top as well as the inherent uncertainty surrounding which individuals for whom the intervention might be most effective, our design distributed interventions across these tiers (in contrast with the high risk tier in which we focus interventions at the top). In the middle risk tier, 40 individuals were selected at random from throughout the tier in the trial group for outreach, and, in the low risk tier, 20 individuals were selected at random from throughout the tier. 
        % Focusing fewer outreach attempts on this lowest-risk tier reflected our prior that, given that relatively few of these individuals returned to jail at baseline, it was less likely that the largest program effects would be observed in this tier.
        
        \item The trial focused more on outreach attempts in the top two tiers because the lower baseline reincarceration risk in the bottom tier suggests the impact in this tier would have to be significantly higher to have an appreciable aggregate impact on reincarceration rates. We expected that it would be easier to statistically detect the impact effect for the top two tiers than for the bottom tier, for two reasons: 1) larger sample size within the same trial length, and 2) larger expected impact effect size for the top two tiers. 
        
        \item This selection process was subject to three constraints:
        
        \begin{enumerate}
            \item To address the fact that people were likely move between risk tiers as the study progresses, individuals were “fixed” to the tier at which they first appeared in the selected 40 or 20 
            
            \item Individuals who have already been included on a model-based list (in either the trial or control group) for two months were excluded from further lists for one year following their second appearance
            
            \item Individuals who have already been contacted by Johnson County’s MCRT within the preceding two months (regardless of whether this outreach was based on the model or for other reasons) were excluded as well (note that this exclusion was applied to both the trial and control groups as well)
        \end{enumerate}
    \end{enumerate}
    
    \item In the control group, the same selection logic as used for the trial group was applied each month to identify the equivalent comparison group for the study.
\end{enumerate}

Although the use of randomization may present concerns about withholding resources from populations in need, we believe those concerns are not present in the case of the current trial for several reasons: First, because the proactive outreach here is additive relative to JCMHC’s current efforts, and individuals in the proposed control group for the purposes of this trial will still be eligible to receive outreach through existing avenues (such as the BJMHS), care and resources were not being withheld from people who would receive them otherwise. Second, resources that could benefit the community are not being artificially withheld for the purposes of measurement: the MCRT can only conduct a limited number of additional outreach attempts given its available resources and this trial design makes use that entire capacity. And, third, it is not clear who would benefit most from the additional outreach (e.g., perhaps individuals with moderate reincarceration risk are more responsive to care than those at the highest risk), so given the limited resources available, it is important to evaluate the impact of outreach across different groups to understand how to have the most impact with these limited resources going forward.

\subsection{Balance and Sample Sizes}
Because new individuals can meet the inclusion criteria during the course of the field trial, randomization was performed dynamically and the sample sizes and balance checks shown in Table \ref{tab:trial_balance} reflect the individuals who were included in either the control or trial arm between May 2019 and January 2020. In total, 3,221 individuals were included in the field trial: 1,009 from Tier 1, 1,475 from Tier 2, and 737 from Tier 3. Although an equal number of individuals were included in the outreach (and control) lists each month from tiers 1 and 2, because a given individual could appear on these lists up to twice and such repeats occurred more frequently in tier 1, a smaller number of unique individuals were included from tier 1 relative to tier 2. For individuals who did appear more than once, their tier assignment and 1 year window for observing outcomes are based on their first occurrence on the respective trial or control list.

Table \ref{tab:trial_balance} also presents balance checks for the trial assignments in each tier, showing p-values from regressing an indicator for trial group assignment against several characteristics important to the outcomes of interest, including sex, age (current and as of first jail booking), prior interactions with the criminal justice system (jail bookings and days in jail), and mental health needs and utilization. Overall, the trial in each tier was well-randomized: of the 24 tests performed here, only one meets a marginal level of significance ($p=0.07$ for the difference in prior JCMHC service utilization between the tier 2 control and trial groups).

\subsection{Outcomes of Interest}
The new outreach program is concerned with two types of outcomes: First, more proximal impacts on mental health needs. And, second, impact on subsequent interactions with the criminal justice system. Although we do not have a direct measurement of crises resolved or averted, we look here at utilization as a signal of mental health needs being met, specifically:

\begin{itemize}
    \item \textbf{JCMHC Services:} Including therapy, psychological evaluation, prescription of medication, rehab, peer support groups, case management, crisis intervention, and other interactions with JCMHC. Although the mental health center tracks outreach, consultations, and paperwork in their services data, we focused here on services identified by JCMHC as reflecting mental health care.

    \item \textbf{JCMHC Admissions:} Admissions reflect intake into JCMHC programs (not necessarily admissions into inpatient psychiatric care such as in a hospital-based setting), potentially indicating a more structured course of care.
\end{itemize}

One limitation to note here is the fact that we only have information on utilization of care offered by JCMHC and not other mental health providers, providing only a limited view of how well mental health needs in the population of interest are addressed. In particular, JCMHC sometimes refers individuals they interact with to other providers and services in the community, and these referrals are unfortunately not well-identified in the data.

To assess the impact on interactions with the criminal justice system, we looked at three outcomes:

\begin{itemize}
    \item \textbf{Arrests:} Arrest data was shared by police departments in Olathe and Overland Park, the two largest communities in Johnson County (together comprising 54\% of the county's population). These data cover only ``on view'' arrests, but include both arrests resulting in a jail booking and notices to appear in court in which the individual is released without being incarcerated, providing an additional information on law enforcement involvement beyond the county-wide jail booking data. Although data was only available from the police departments in these two communities, the trial groups were well-balanced for residence addresses in these communities (57.5\% of control group individuals vs 57.4\% of those receiving outreach overall; by tier, the difference in residence rates between control and outreach groups was: 0.9pp in tier 1, -1.3pp in tier 2, and 0.6pp in tier 3).

    \item \textbf{Jail Bookings:} We included as outcomes any new jail bookings associated with ``on view'' arrests, arrest warrants, or bench warrants. Although bench warrants, which are generally issued due to a failure to appear in court for a case related to a prior arrest, the Johnson County staff was interested in including these bookings in the analysis because they may in many cases be readily preventable links in the cycle of incarceration.

    \item \textbf{Days in Jail:} As an additional indication of the extent of involvement with the criminal justice system, we looked at the number of days individuals spent incarcerated in Johnson County jail. One limitation here, however, is that we did not have access to prison data from the Kansas Department of Corrections (or jail data from nearby counties), so these results only reflect time spent incarcerated in Johnson County jail while awaiting trial or serving relatively short sentences, but not time spent serving longer sentences in prison.
\end{itemize}

\begin{table}
 \caption{Sample sizes (unique individuals) and balance checks for each trial tier (p-value for difference between trial and control arms)}
  \centering
  \begin{tabular}{lccc}
    \toprule
         & Tier 1     & Tier 2 & Tier 3 \\
    \midrule
    Control N               & 232 & 339 & 168 \\
    Trial N                 & 276 & 333 & 175 \\
    \midrule
    Sex                     & 0.92 & 0.12 & 0.93  \\
    Current age             & 0.14 & 0.18 & 0.84  \\
    Age at first booking    & 0.16 & 0.97 & 0.78  \\
    Jail bookings           & 0.53 & 0.51 & 0.55  \\
    Days in jail            & 0.17 & 0.37 & 0.41  \\
    JCMHC admissions        & 0.73 & 0.66 & 0.73  \\
    JCMHC services          & 0.93 & 0.07 & 0.42  \\
    Mental health flags     & 0.47 & 0.15 & 0.79  \\
    \bottomrule
  \end{tabular}
  \label{tab:trial_balance}
\end{table}

\section{Field Trial Results}

\subsection{Model Generalization Performance}
\begin{figure}
  \centering
  \includegraphics[width=0.5\linewidth]{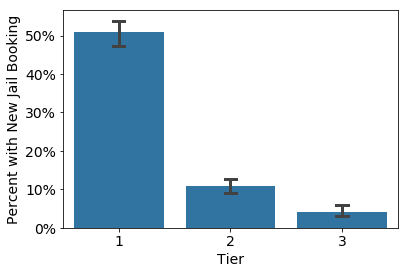}
  \caption{New jail bookings by score tier in the trial control group. Error bars reflect standard error of the mean.}
  \label{fig:booking_tier}
\end{figure}

An initial aim of the field trial was to confirm the predictive performance of the model on future data. To evaluate this generalization performance, we measured the jail booking outcomes among the individuals in the control group in each tier. These results, shown in Figure \ref{fig:booking_tier}, confirm the model's ability to predict future jail incarceration. The relative jail booking rates stacks as expected across the three tiers and, in particular, individuals in tier 1 are at much higher risk of a subsequent booking, with more than half of these individuals experiencing a subsequent booking (similar to the validation performance of the top model vigintile in Figure \ref{fig:model_results}B. The exaggerated performance of tier 1 relative to tiers 2 and 3 is also likely due to the different selection strategies between the tiers (with tier 1 focusing on the highest risk scores in the tier while individuals were randomly selected for inclusion in the other tiers, as described in the trial design above).

\subsection{Mental Health Utilization Outcomes}
The primary aim of the field trial was to understand the extent to which the proactive outreach program could identify and serve a population both with high needs for mental health intervention and for whom addressing those needs might reduce their risks of future criminal justice involvement. Ideally, to assess the more immediate impacts of the program, we would like to have some direct measure of the actual mental health needs of the population, but lacking this information, we focused on utilization of JCMHC services and programs as a proxy for needs being met. 

\begin{figure}
  % \centering
  \includegraphics[width=0.3\linewidth]{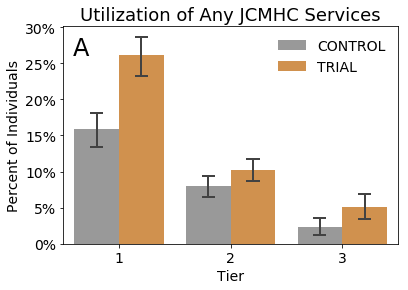}
  \includegraphics[width=0.3\linewidth]{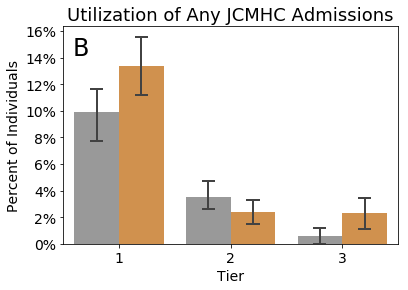}
  \includegraphics[width=0.3\linewidth]{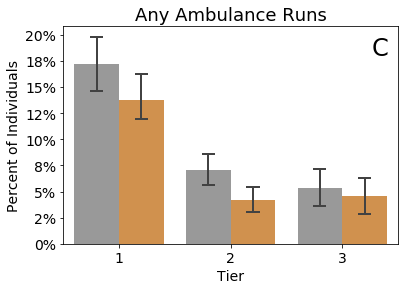} \\
  \includegraphics[width=0.3\linewidth]{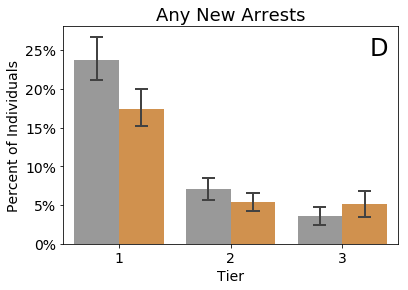}
  \includegraphics[width=0.3\linewidth]{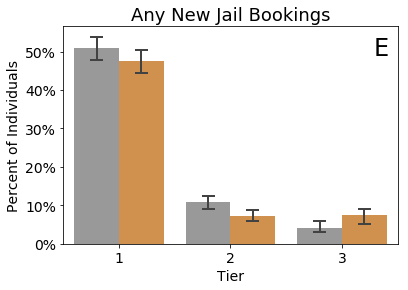}
  \includegraphics[width=0.3\linewidth]{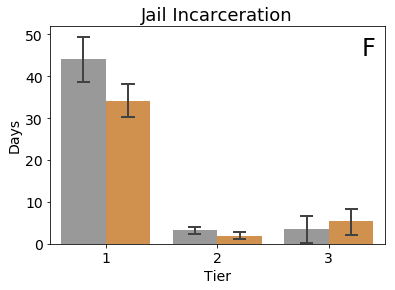}
  \caption{Outcome measures for trial and control groups. Error bars reflect standard error of the mean.}
  \label{fig:trial_results}
\end{figure}

Baseline levels of JCMHC utilization in the control group tracked strongly with modeled risk of reincarceration (Figure \ref{fig:trial_results}(A-B), gray bars). While the models weren't trained to detect mental health needs directly, this relationship is not surprising given the well-documented concentration of mental illness in the incarcerated population \cite{Sarteschi2013MentallySynthesis,Kennedy-Hendricks2016ImprovingIllness,James2006MentalInmates}. Notably, there is some degree of endogeneity here as well, because JCMHC has several programs specifically focused on serving individuals in or recently released from jail. In the group receiving outreach, utilization of JCMHC services was increased across all three model score tiers (Figure \ref{fig:trial_results}A, orange bars), but with particularly large effects among the highest-risk individuals. Likewise, admissions into formal programs of care was elevated among tier 1 individuals receiving outreach, while effects among individuals at lower risk of reincarceration were small and inconsistent.

Table \ref{tab:trial_results} shows detailed results for individuals in tier 1, with effect sizes and significance levels estimated using ordinary least squares (OLS) regressions controlling for demographics, prior interactions with the criminal justice system, and mental health utilization. Use of JCMHC services among tier 1 individuals receiving outreach was significantly elevated relative to individuals in the control group, by 10.8 percentage points (against a baseline of 16\% in the control group, p=0.002). Admissions to JCMHC programs were also elevated among individuals receiving proactive outreach in tier 1, increasing by 4.5pp from a baseline of 10\% in the control group, although this result only reaches a moderate level of statistical significance (p=0.10).

Although the data reflects only interactions with JCMHC and therefore doesn't provide a comprehensive picture of all the ways in which individuals' needs may be getting met (for instance, through other community-based or private mental health providers), we nevertheless believe there is a considerable degree of unmet mental health need in the population the MCRT outreach program serves. Likewise, in terms of measurement, high levels of service through other providers would be expected to lower the baseline level of need and tend towards decreasing the measurable impact of the current outreach. The lack of data from other providers isn't a source of concern that we might over-estimate the impact of this program by focusing on JCMHC data -- in fact, because MCRT may sometimes refer individuals to other providers (and these referrals are not well-tracked in the available data), the level of overall mental health service utilization as a result of the outreach may be somewhat under-estimated here.

As an additional indication of outcomes related to mental health needs, we also looked at ambulance runs involving the individuals in the trial. While many ambulance runs will be in response to emergencies unrelated to mental health issues, some calls may result from individuals in crisis, struggling with substance abuse, or suffering from severe mental illness. To the extent that some of these crises and other emergencies might have been averted by counseling, rehabilitation, or other JCMHC services, we might expect a reduction in ambulance calls involving individuals receiving outreach than similarly-situated individuals in the control group. The results in Figure \ref{fig:trial_results}C are suggestive that this may be the case -- across all three tiers, fewer individuals were involved in ambulance runs in the group receiving outreach relative to the control group. This effect was particularly large in tiers 1 and 2, a decrease of around 3pp in each case. Although the tier 1 result itself falls well short of statistical significance (Table \ref{tab:trial_results}, p=0.37), pooling the data from tiers 1 and 2 yields an estimated effect size of -2.7pp and much closer to traditional significance levels (p=0.12), lending further suggestive evidence of the ability of the proactive outreach to meet the mental health needs of the individuals it serves, particularly when well-targeted to individuals with high needs.

\begin{table}
 \caption{Estimated outreach effects for outcome measures on tier 1 individuals. Effect sizes, standard errors, and p-values calculated with robust OLS regression controlling for prior jail bookings, days in jail, mental health service utilization, age at first booking, current age, and sex.}
  \centering
  \begin{tabular}{lcccc}
    \toprule
    Outcome     & Control Group Mean & Outreach Effect & Std Error & p-value \\
    \midrule
    Any JCMHC Service & 16\% & +10.8 pp & 3.5 pp & 0.00 \\
    Any JCMHC Admission & 10\% & +4.5 pp & 2.8 pp & 0.10 \\
    Any Ambulance Run & 17\% & -2.9 pp & 3.2 pp & 0.37 \\
    Any Arrest & 24\% & -6.0 pp & 3.7 pp & 0.10 \\
    Any Jail Booking & 51\% & -3.7 pp & 4.5 pp & 0.41  \\
    Days in Jail & 44.1 days & -11.8 days & 6.8 days & 0.08 \\
    \bottomrule
  \end{tabular}
  \label{tab:trial_results}
\end{table}

\subsection{Criminal Justice Outcomes}
The broader aim of the new outreach program is to break the cycle of incarceration by addressing mental health needs of individuals at risk of facing criminal justice involvement in the near future. Unlike mental health needs, we do have more direct measures of several criminal justice outcomes, including subsequent arrests (using police data from the two largest communities in the county) as well as new bookings and time spent in the Johnson County jail system. Figure \ref{fig:trial_results}(D-F) shows the outcomes for both trial groups across the three criminal justice measures. The baseline rates in the control group are much higher in tier 1 than tiers 2 and 3, consistent with the model's predictive power discussed above, and outreach effects are generally small in magnitude and inconsistent in direction in the lower two score tiers. As such, we focused our analysis on the results for individuals in tier 1, with detailed outcomes and estimated effect sizes for this tier shown in Table \ref{tab:trial_results} using the regression controls described above.

The fraction of individuals in each group who were arrested in Olathe or Overland Park in the year following their inclusion in the trial is shown in Figure \ref{fig:trial_results}D. As noted above, arrest data was only available from police departments in these two cities, though they comprise the majority of the county's population and include arrests of both residents and non-residents alike, making them still of interest as an outcome variable despite their limited scope. Moreover, because the arrest data include both arrests that result in a jail booking and those which result in a notice to appear in court without being booked, they may provide a more complete picture of criminal justice involvement in that regard. While 24\% of control group individuals in tier 1 of the risk score experienced a subsequent arrest, those receiving proactive mental health outreach saw a considerable drop in their subsequent arrest rate, with an estimated effect size of -6.0pp at a moderate level of statistical significance (p=0.10). The impact on new jail bookings, however, was less clear: although directionally consistent with the arrest data at an estimated 3.7pp decrease for the outreach group, this result fell far short of statistical significance (p=0.41). In addition to statistical variation, two factors might contribute to the difference between the arrest and booking results: First, because many of the subsequent bookings are for bench warrants (generally due to a failure to appear in court), these may contribute to the high baseline booking level in this group and increase the variance of the result. Second, because the arrest data includes "notice to appear" arrests aside from bookings, these likely reflect less severe circumstances which may in turn be more readily preventable through mental health interventions.

Because the number of days an individual spends in jail likely also serves as a reflection of the degree of severity of the circumstances surrounding their incarceration, we looked at this as another measure of criminal justice outcomes (Figure \ref{fig:trial_results}F). Consistent with a decrease in both arrests and new bookings, individuals in the top tier of the risk score who received outreach spent nearly 12 fewer days in jail relative to the control group mean of 44 days (p=0.08). Although the criminal justice outcomes here generally reach only moderate levels of statistical significance, they tell a consistent story across outcomes and are suggestive that the impact of the mental health interventions could be considerable in practical terms. 

\section{Impact of COVID-19}

One challenge encountered during the course of this trial was the onset of the COVID-19 pandemic in early 2020, dramatically upending social services and law enforcement practices. As a result, we suspected that outreach impacts may not be consistent prior to and during the pandemic, which may have an impact on both the overall estimates and statistical power of the analysis discussed above.

\begin{figure}
  \centering
  \includegraphics[width=0.4\linewidth]{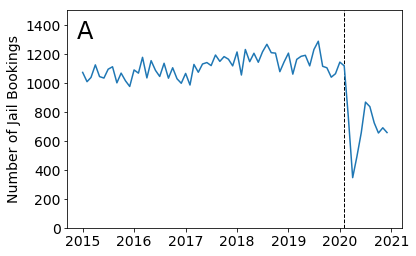}
  \includegraphics[width=0.4\linewidth]{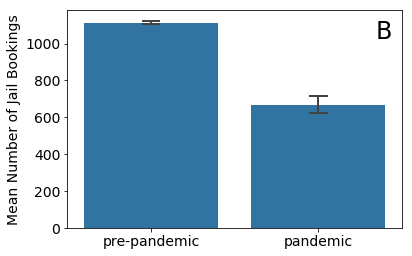}
  \caption{(A) Number of Johnson County jail bookings by month since 2015, with the dashed line indicating the last month prior to the pandemic (Feb 2020). (B) Data from (A) aggregated to the mean number of monthly bookings before and during the pandemic. Error bars reflect standard error of the mean.}
  \label{fig:booking_covid}
\end{figure}

For example, total number of bookings into the Johnson County jail was relatively stable over time prior to the pandemic, with around 1,100 bookings per month (Figure \ref{fig:booking_covid}), but dropped dramatically at the outset of the pandemic. Despite rebounding somewhat after the first few months, bookings remain significantly below pre-pandemic levels (on average, monthly bookings during the pandemic are about 40\% below their previous levels). Furthermore, it seems reasonable to hypothesize that this decrease in jail bookings isn't uniformly distributed across type or severity of criminal conduct. If it were the case that reductions in enforcement and incarceration for less-severe conduct was driving this decrease, to the extent that these are also circumstances that are most readily preventable through the mental health outreach explored here, the pandemic could have an appreciable implications for the impact of the outreach here. 

Likewise, the nature of the MCRT outreach changed during the pandemic: initially, outreach attempts were first a phone call followed by a door knock, but due to safety concerns during the pandemic, the second attempt was changed to a direct mail piece instead. While this change preserves the use of two separate pieces of contact information for the two attempts (for an individual's phone number on record was out of date, but their address wasn't), it certainly reflects a less personal, lower-touch contact than the pre-pandemic outreach program.

To explore the impact of the COVID-19 pandemic on the outreach program, we compared results for individuals who were included in the trial prior to October 2019 to those included in the trial later, focusing as above on tier 1 of the reincarceration risk score. Although all individuals in the trial had some portion of their 1-year follow-up window overlap with the pandemic, this division allowed us to compare results for individuals with at least 6 months of pre-pandemic outcomes with those for whom the majority of their follow-up window was during the pandemic. 

The results shown in Table \ref{tab:trial_covid} split out these pre- and during-pandemic estimates of outreach effects for tier 1 individuals using the same regression controls as in Table \ref{tab:trial_results}. Although the analysis is generally under-powered, the trends suggest the impact of outreach on JCMHC utilization is relatively similar before and during the pandemic (somewhat surprisingly, admissions to JCMHC program appear to be higher during the pandemic), suggesting the mental health center was still able to reach individuals with need despite the changes to the outreach modalities during the pandemic (notably the stress of the pandemic itself may also have increased the levels of need for mental health services or receptiveness to seeking help \cite{Pfefferbaum2020MentalPandemic,Panchal2021TheUse}). For all other outcomes of interest, however, the estimated effect sizes are larger pre-pandemic.\footnote{Notably, as above, estimated effect sizes on ambulance runs are similar for individuals in tiers 1 and 2 pre-pandemic. Pooling these individuals in the analysis yields a statistically significant estimated effect on ambulance runs of -4.9pp (p=0.03).} While not fully conclusive due to the limited power, this result is consistent with the hypothesis that the changes to criminal justice practices during the pandemic might serve to limit the effects of outreach and suggest that the disruptive and wide-spread changes during the COVID-19 pandemic may be a limiting factor in the present trial's power.

\begin{table}
 \caption{Estimated outreach effects for outcome measures on tier 1 individuals, disaggregated by individuals with at least 6 months of outcomes prior to the COVID-19 pandemic (those included in the trial prior to October 2019) and those with outcome windows predominantly during the pandemic. Effect sizes, standard errors, and p-values calculated with robust OLS regression controlling for prior jail bookings, days in jail, mental health service utilization, age at first booking, current age, and sex.}
  \centering
  \begin{tabular}{lcccc}
    \toprule
    &     \multicolumn{2}{c}{Pre-Pandemic} & \multicolumn{2}{c}{During Pandemic} \\
    \cmidrule(r){2-3} \cmidrule(r){4-5}
    Outcome     & Outreach Effect & p-value & Outreach Effect & p-value \\
    \midrule
    Any JCMHC Service   & +9.8 pp & 0.03 & +11.3 pp & 0.05 \\
    Any JCMHC Admission & +0.5 pp & 0.89 & +8.3 pp & 0.10 \\
    Any Ambulance Run & -5.8 pp & 0.17 & +2.5 pp & 0.66 \\
    Any Arrest          & -7.3 pp & 0.14 & -3.7 pp & 0.50 \\
    Any Jail Booking    & -5.4 pp & 0.37 & -1.7 pp & 0.82  \\
    Days in Jail        & -11.0 days & 0.24 & -9.4 days & 0.39 \\
    \bottomrule
  \end{tabular}
  \label{tab:trial_covid}
\end{table}

\section{Discussion}

The analysis of pandemic effects above highlights a challenge faced by many organizations, both private and public. With considerable lack of clarity in what ``normal'' will look like even as the pandemic itself recedes, planning for even the near future involves a considerable amount of uncertainty. Will some of the changes to law enforcement and prosecution that were implemented over the course of the past year take hold more permanently? What longer-term impacts will the pandemic have on the mental health and substance dependence of the population? What other changes to society or social services will persist beyond the pandemic? These questions pose an important threat to the external validity to any impact evaluation conducted before or during the pandemic, including the work described here. Ongoing efforts to understand how the changing policy and social context effects these results will no doubt be important as we move beyond the pandemic.

One of the key questions the field trial sought to address was where in the reincarceration risk score distribution the outreach treatment might be most effective. In the context of the current field trial, the largest and most consistent effects were found when targeting outreach to the highest-risk individuals (tier 1). Nevertheless, there is also considerable room for continued refinement and investigation here. The design of the present field trial focusing on three risk-score tiers is an important limitation of the present work and inherently reflects the resource constraints of the mental health center doing the outreach. Ideally a large enough trial could randomize across the population without regard to the score and explore treatment effects across the modeled risk in a post hoc fashion or even use the trial results as training data to build a heterogeneous treatment effects model directly. 

Because the MCRT only had capacity to do outreach to 100 individuals each month, we instead focused on three potential outreach targeting strategies represented by the three score tiers, with an initial hypothesis that the level of need of the highest-risk individuals might be too high for the program to have an appreciable effect and larger impacts would be observed for individuals with more moderate risk levels. Nevertheless, the considerable drop in baseline reincarceration rates from tier 1 to the other two tiers (Figure \ref{fig:booking_tier}) suggests the tiers could have been designed with a more gradual decline in risk across groups. Exploring variations in treatment effects among higher-risk individuals will be an important direction for future work.

If future analyses do find outreach to be more effective at a more intermediate score range, one important consideration for implementation will be the interaction between process fairness and efficiency. That is, to what extent should efforts be devoted to understanding needs among the highest-risk individuals and designing new interventions that may be more appropriate for them versus helping as many people as possible who are likely to respond to the existing intervention? Likewise, how might that decision differentially impact individuals across salient attributes such as race, gender, or age. Although these process questions become less relevant when targeting focuses on the highest-risk individuals (as the present results suggest), there are nonetheless important considerations around the fairness of the model's recommendations and the intervention's effectiveness across groups. We have written separately about algorithmic fairness in this \cite{Rodolfa2020MachineFair} and related \cite{Rodolfa2020CaseInterventions} contexts when focusing on the top of the score, and will be implementing these methods to ensure the fairness of the system as we work towards deployment for ongoing use.

% \section{Ethical considerations}
% \rgcomment{we should discuss 1)ethical issues 2) community involvement and 3) other related things here or possibly earlier}
\section{Conclusions}
Although the statistical power of this trial was limited, particularly with regards to impacts on criminal justice outcomes, it is also worth noting that JCMHC didn't hire additional staff for this new effort, so the marginal cost of implementing the program is effectively the opportunity cost of other activities the staff could have been doing in this time. Because there is little formal evidence of the relative impacts of these other activities, it is difficult to fully understand the relative costs and benefits of the proactive outreach. Nevertheless, in the absence of other evidence, even the moderately significant and suggestive results reported here certainly provide a promising indication that should be followed up with continued efforts and further evaluation, especially as society begins to move back to a pre-pandemic state of normalcy.

The proactive outreach described here is considerably low-touch compared to interventions such as FACT or mental health courts, and it seems encouraging that merely offering mental health resources can improve outcomes in this vulnerable population. Crucially, we don't envision such outreach as a replacement for these other interventions, but rather one of many tools that can operate concurrently at different intervention points. We believe the proactive nature of the program here is of particular interest and makes it well-suited to supplement existing, reactive efforts to connect individuals to mental health care (or other social services) at the time of criminal justice involvement. One general finding of this work that may be of interest to researchers and policy makers is the importance of focusing the MCRT's limited resources on individuals for whom outreach might be most effective. Here, this was accomplished using predictive modeling to identify individuals at higher risk for future jail incarceration, an approach which could readily be applied in many other jurisdictions. The concentration of treatment effects among this highest-risk population is also notable in this regard, and we are not aware of other randomized evaluations of targeted outreach aiming to break the cycle of incarceration. Likewise, it may be possible that, even where some other interventions have shown mixed results \cite{Kim2015TheSystem}, a subset of the population potentially receptive to such interventions could be identified with similar targeting in these contexts as well.

While much work remains to fully implement, monitor, and improve this system in Johnson County, we hope that the case study presented here can prove informative to the design and evaluation of intervention models to proactively connect social and mental health workers with individuals in need of care to avoid incarceration in other jurisdictions as well.

% In sum, this work provides a framework and prototype system for Johnson County as well as many other jurisdictions that are part of the Data Driven Justice Initiative as they develop intervention models to proactively connect social and mental health workers with individuals in need of care to avoid incarceration.

\section{Acknowledgements}
In addition to our funders, \krcomment{TODO: add funders}, we would like to thank the fellows and mentors of the Data Science for Social Good Fellowhip who worked on the initial phases of this project as well as the many individuals in Johnson County, KS, who were instrumental in making it happen, including the County Manager's Office, Mental Health Center Leadership and Caseworkers, Department of Corrections, Office of the Criminal Justice Coordinator, and the many agencies than contributed data to support the work.

\section{Abbreviations}%% if any
AUC - Area Under the Curve \\
BJMHS - Brief Jail Mental Health Screen \\
DDJ - Data-Driven Justice \\
DTI - Department of Technology and Innovation \\
EMS - Emergency Medical Services \\
FACT - Forensic Assertive Community Treatment \\
ITT - Intent-To-Treat \\
JCMHC - Johnson County Mental Health Center \\
LSI-R - Level of Service Inventory - Revised \\
MCRT - Mobile Crisis Response Team \\
ML - Machine Learning \\
ROC - Receiver Operating Characteristic

\section{Availability of data and materials}%% if any
Data for this work was shared through a partnership and data use agreement with the county government of Johnson County, KS (which collected and made available data from the county- and city-level agencies in their jurisdiction as described in the Methods above). Although the sensitive nature of the data used here required that the work was performed under a strict DUA and the data cannot be made publicly available, researchers or practitioners interested in collaborating on these projects or with the agencies involved should contact the corresponding author (rayid@cmu.edu) for more information and introductions. The code used here for building and evaluating predictive models is available at github.com/dssg/triage.

\section{Ethics approval and consent to participate}%% if any
This study was reviewed and approved by the Carnegie Mellon University Office of Research Integrity and Compliance's IRB (Protocol Number: STUDY2021\_00000332).

\section{Competing interests}
The authors declare that they have no competing interests.

% \section*{Consent for publication}%% if any
% Not applicable

\section{Authors' contributions}
KTR and ES built and evaluated machine learning models, developed methodology and software for the project, and coordinated with county staff on implementation. 
KTR additionally analyzed the results of the field trial and wrote the original draft of the manuscript. 
JY implemented and ran the models to general outreach lists as well as assisted with analysis of results.
SY gathered data from county agencies, developed record linkage software to combine them, and de-identified and exported the linked data for analysis.
RS, KM, AD, RM, and BS coordinated data acquisition, outreach operations, and provided supervision and support for model development and implementation.
CS provided project management support and coordination with stakeholders in Johnson County.
CH assisted with model development and evaluation.
RG scoped and conceptualized the work, provided research supervision, and aquired funding.
All authors read and approved the final manuscript.

\bibliographystyle{plain}
\bibliography{kit_mendeley_refs}

\end{document}